\documentclass[letterpaper]{article} 
\usepackage{aaai2026}  
\usepackage{times}  
\usepackage{helvet}  
\usepackage{courier}  
\usepackage[hyphens]{url}  
\usepackage{graphicx} 
\usepackage{natbib}  
\usepackage{caption} 
\usepackage{algorithm}
\usepackage{algorithmic}
\usepackage{xcolor}
\usepackage{multirow}
\usepackage{amsmath}
\usepackage{amsthm}
\newtheorem{definition}{Definition}

\usepackage{newfloat}
\usepackage{listings}
\DeclareCaptionStyle{ruled}{labelfont=normalfont,labelsep=colon,strut=off} 
\floatstyle{ruled}
\newfloat{listing}{tb}{lst}{}
\floatname{listing}{Listing}
\usepackage{tabularx}
\usepackage{booktabs}

\usepackage{multirow}
\usepackage{amsthm}
\usepackage{subfigure}
\usepackage{bm}
\usepackage{amssymb}

\title{ResMAS: Resilience Optimization in LLM-based Multi-agent Systems}
\author{
    Zhilun Zhou\textsuperscript{\rm 1},
    Zihan Liu\textsuperscript{\rm 1},
    Jiahe Liu\textsuperscript{\rm 1},
    Qingyu Shao\textsuperscript{\rm 1},
    Yihan Wang\textsuperscript{\rm 1},\\
    Kun Shao\textsuperscript{\rm 2},
    Depeng Jin\textsuperscript{\rm 1},
    Fengli Xu\textsuperscript{\rm 1}\thanks{Corresponding author: fenglixu@tsinghua.edu.cn}
}
\affiliations{
    \textsuperscript{\rm 1}Department of Electronic Engineering, BNRist, Tsinghua University\\
    \textsuperscript{\rm 2}Huawei Noah’s Ark Lab\\
    \{zzl22,zh-l23,liujiahe22,shaoqy23,wangyiha23\}@mails.tsinghua.edu.cn,\\
    shaokun2@huawei.com,
    \{jindp,fenglixu\}@tsinghua.edu.cn

}

\usepackage{bibentry}

\begin{document}

\maketitle

\begin{abstract}
    Large Language Model-based Multi-Agent Systems (LLM-based MAS), where multiple LLM agents collaborate to solve complex tasks, have shown impressive performance in many areas. However, MAS are typically distributed across different devices or environments, making them vulnerable to perturbations such as agent failures. While existing works have studied the adversarial attacks and corresponding defense strategies, they mainly focus on reactively detecting and mitigating attacks after they occur rather than proactively designing inherently resilient systems.
    In this work, we study the resilience of LLM-based MAS under perturbations and find that both the communication topology and prompt design significantly influence system resilience. Motivated by these findings, we propose ResMAS: a two-stage framework for enhancing MAS resilience. First, we train a reward model to predict the MAS’s resilience, based on which we train a topology generator to automatically design resilient topology for specific tasks through reinforcement learning. Second, we introduce a topology-aware prompt optimization method that refines each agent’s prompt based on its connections and interactions with other agents. 
    Extensive experiments across a range of tasks show that our approach substantially improves MAS resilience under various constraints. Moreover, our framework demonstrates strong generalization ability to new tasks and models, highlighting its potential for building resilient MASs.
\end{abstract}


\begin{links}
    \link{Code}{https://github.com/tsinghua-fib-lab/ResMAS}
\end{links}

\section{Introduction}

Large language model-based multi-agent systems (LLM-based MAS; hereafter referred to as \textbf{MAS}) have exhibited impressive capabilities across a wide range of domains, including software development~\cite{qian2024chatdev,hong2023metagpt}, mathematical problem solving~\cite{li2023camel}, and operations research~\cite{xiao2023chain}. In an MAS, multiple LLM agents collaborate to solve complex tasks through certain role-play and communication patterns, often achieving superior performance compared to single-agent approaches.
Despite these advances, agents within an MAS are typically deployed across different devices or environments, making the MAS vulnerable to perturbations and attacks. As a result, the resilience, defined as the ability of a system to maintain its functionality under perturbations~\cite{cohen2000resilience,gao2016universal,ding2024comprehensive}, has become an important issue for MAS. 

In recent years, there have been extensive studies about resilience of MAS under adversarial attacks. Several studies propose attack strategies such as prompt injection~\cite{yu2024netsafe,huangresilience} and knowledge editing~\cite{ju2025investigating,ju2024flooding}, and find that harmful information can rapidly spread in MAS~\cite{gu2024agent,lee2024prompt}. Other works focus on detecting and defending against attacks using techniques like psychological testing~\cite {zhang2024psysafe} and graph-based anomaly detection~\cite{wang2025g}.
However, existing studies mainly focus on the safety issue of MAS caused by adversarial attacks, while largely overlooking more common, non-malicious errors in MAS such as failures or miscommunication. Furthermore, prior works concentrated on detecting and mitigating attacks, with little focus on proactively designing resilient MAS structures that can better tolerate errors.

In this work, we study the resilience of MAS under random agent failures. We find that MAS exhibits significantly higher resilience than a single agent, and the resilience increases with the number of agents and communication links. Moreover, we find that both the topology and prompts of an MAS have a strong impact on its resilience.
Motivated by these, we propose a two-stage framework to automatically optimize the topology and prompts to improve the resilience of MAS, named \textbf{ResMAS}. 
In the first stage, we fine-tune an open-source LLM serve as a topology generator capable of producing resilient MAS structures. Specifically, to avoid the high time cost of evaluating an MAS's resilience, we train a GNN-based reward model to predict the resilience of MAS for various tasks. This reward model is then used to fine-tune the LLM using the Group Relative Policy Optimization (GRPO)~\cite{shao2024deepseekmath} algorithm, enabling it to automatically generate resilient MAS topologies.
In the second stage, we propose a topology-aware prompt optimization method to optimize the system prompt for each agent in the MAS based on the generated topology. 
Specifically, we apply the MAS to a training set of the target task, and refine the system prompt for each agent based on the prompts of its neighbors as well as its interaction history with neighbors.

We conduct experiments on three benchmarks, including commonsense reasoning, math, and game. The results demonstrate that our framework can generate more resilient MAS compared to existing approaches for prompt and topology optimization. Moreover, we highlight the versatility of our framework by applying it to optimize the accuracy of MAS on certain tasks, where it achieves the Pareto-optimal performance compared with baselines. In addition, our framework shows strong generalization ability to unseen tasks such as code generation, and to new agents built upon different backbone models.

In summary, our contributions are as follows:
\begin{itemize}
    \item We systematically study the resilience of MAS under random agent failures, revealing that both the topology and agent prompts have significant influence on system resilience.
    \item We propose ResMAS: a two-stage framework that automatically optimizes the topology and agent prompts of MAS to enhance resilience. Specifically, we fine-tune an LLM to generate robust topologies, and further refine the prompts in a topology-aware manner.
    \item Extensive experiments on various tasks show that our framework is able to generate MAS with superior resilience and demonstrates strong generalization ability across tasks and models.
\end{itemize}

\section{Related Work}
\label{sec:related work}

\subsection{Resilience of MAS}
The distributed architecture of MASs inherently increases their susceptibility to perturbations. Prior research has focused on adversarial attack and defense methods for MAS. 
Specifically, some studies propose attack methods such as prompt injection~\cite{yu2024netsafe,huangresilience,tian2023evil} and knowledge editing~\cite{ju2025investigating,ju2024flooding}. Additionally, it has been demonstrated that harmful information can spread extremely fast in MAS, where a single jailbreak agent can undermine the whole MAS~\cite{gu2024agent,lee2024prompt}. In response, some studies focus on detecting and defending against attacks through methods like psychological test~\cite{zhang2024psysafe} and graph-based anomaly detection~\cite{wang2025g}.
However, existing studies mainly focus on the safety issue of MAS caused by adversarial attack, but overlook the resilience to more common, non-malicious agent errors. Furthermore, prior works emphasize detecting and mitigating attacks, with little focus on proactively designing resilient MAS structures that can tolerate errors by design.

\subsection{Automated Design of MAS}
The architecture of MAS, including the agents' roles and prompts as well as the communication patterns between agents, significantly affects the performance of MAS on tasks. Early studies often manually design MAS structures for specific tasks, usually based on human collaboration mechanisms~\cite{DBLP:conf/icml/Du00TM24,hong2023metagpt}. Recently, there have been more studies on automated design and optimization of MAS. For instance, Agentverse lets LLM generate and adjust the agent composition based on the status of the task~\cite{chen2023agentverse}. G-designer proposes to optimize the communication network of agents through a variational graph auto-encoder~\cite{zhang2024g}. GPTSwarm represents multi-agent systems as composite graphs and optimizes node-level prompts as well as edges between agents~\cite{zhuge2024gptswarm}.
However, all of the existing methods focus on optimizing the performance of MAS on certain tasks rather than resilience.

\section{Preliminaries}
\label{sec:preliminaries}
\subsection{Motivation}
\label{sec:motivation}

In this study, we use a widely used MAS framework~\cite{du2023improving, wang2025mixtureofagents}, where multiple LLM agents discuss for several rounds to answer a question.
\begin{definition}[\textbf{LLM-based Multi-Agent System}]
Specifically, the LLM agents can be modeled as a directed graph $\mathcal{G}=\{\mathcal{V}, \mathcal{E}, \mathcal{P}\}$, where $\mathcal{V}=\{v_1,v_2,\ldots,v_{N}\}$ is the set of nodes, each node is an LLM agent, $\mathcal{E}$ is the set of edges, and $\mathcal{P}=\{P_1,P_2,\ldots,P_{N}\}$ represents the prompts for agents. We also refer to the graph $\mathcal{G}$  as the \textit{communication topology} of LLM agent groups.
Given a query $q$, each agent $v_i\in \mathcal{V}$ independently generates an initial response $r_i^{(1)}=v_i(q)$. Then in round $t (t\ge 2)$, each agent observes the previous answers of its in-neighbors, and updates its own answer:
\begin{equation}
    r_i^{(t+1)} = v_i(\{r_j^{(t)}|j\in \mathcal{N}_{in}(v_i)\}),
\end{equation}
where $\mathcal{N}_{in}(v_i)$ denotes the in-neighboring nodes of $v_i$.
After $T$ rounds, the final answer is obtained by aggregating the responses of all agents
\begin{equation}
    r^{(T)} = Aggregate(r_1^{(T)},r_2^{(T)},\ldots,r_{N}^{(T)}).
\end{equation}
\end{definition}

In practice, different agents in an MAS are often deployed in a distributed way, i.e., they may be on different devices and environments, making MAS vulnerable to external perturbations. In this work, we define the perturbation as random agent failure, where in each round of the collaboration, each agent independently has a probability $p$ to output a random response. We call the probability $p$ the error rate.

\begin{figure}[h]
\centering
    \subfigure[]{
    {\label{subfig:single_vs_MAS}}
    \includegraphics[width=.45\linewidth]{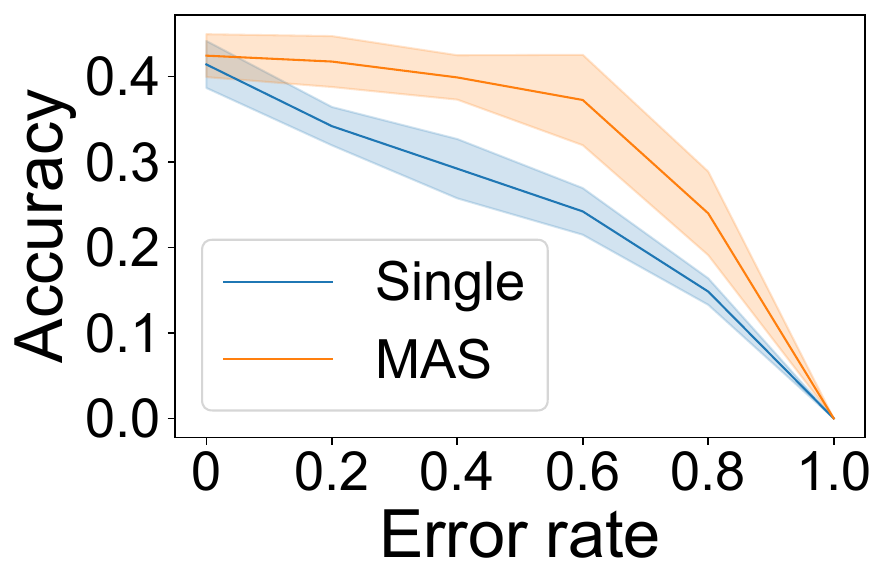}
    }
    \subfigure[]{
    {\label{subfig:resilience_def}}
    \includegraphics[width=.45\linewidth]{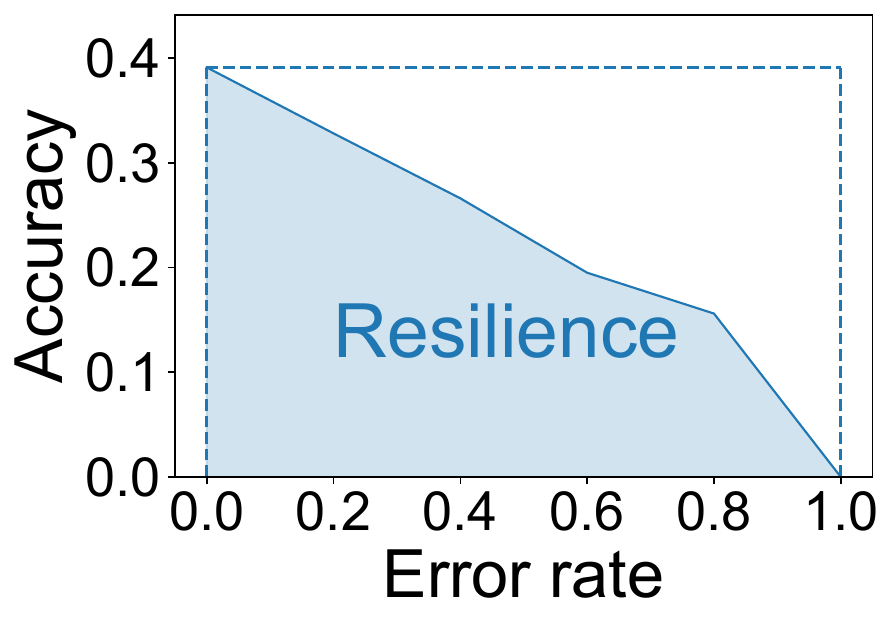}
    }
    \subfigure[]{
    {\label{subfig:agentnum_vs_resilience}}
    \includegraphics[width=.45\linewidth]{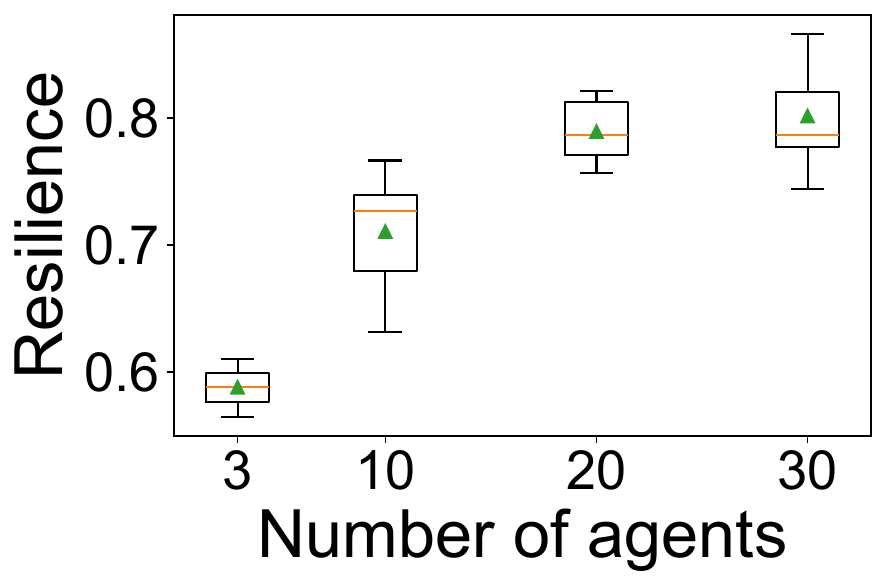}
    }
    \subfigure[]{
    {\label{subfig:edgenum_vs_resilience}}
    \includegraphics[width=.45\linewidth]{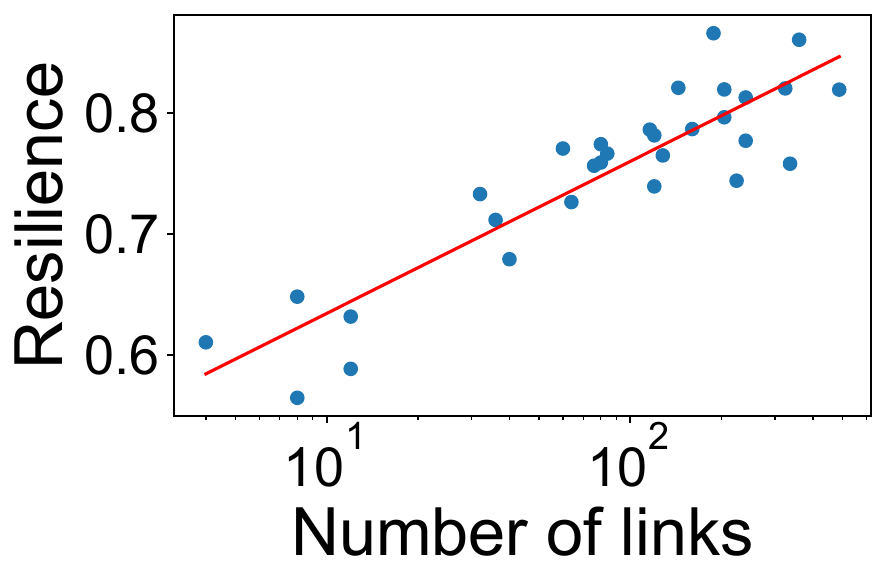}
    }
\caption{(a) Performance of single agent and MAS under perturbation. (b) Definition of resilience. (c)(d) Resilience generally increases with the number of agents and links.}
\label{fig:motivation}
\end{figure}

We first study the behavior of MAS under perturbation. Specifically, we generate 30 random graphs with different agent numbers and edge numbers, and use them as the communication topology to construct 30 MASs. We then test the accuracy of these MASs on the Chess move validity tasks (details in the dataset section) under different error rates ($p=0,0.2,0.4,0.6,0.8,1$). We compare the results of MASs with single agent in Figure~\ref{subfig:single_vs_MAS}. It can be observed that as the error rate increases, both the performance of single agent and MAS drops. However, the performance of MASs remains much higher than that of a single agent under different error rates, indicating that MASs are more resilient to perturbations.

The resilience of a system is defined as its capability to maintain its functionality under perturbations~\cite{cohen2000resilience,gao2016universal}. Generally, the functionality of an MAS is reflected by its performance on certain tasks. Therefore, we quantitatively define the resilience of MAS as follows:

\begin{definition}[\textbf{Resilience of MAS}]
Given an MAS $\mathcal{G}$ and a target task consisting of $K$ problems $\mathcal{T}=\{q_1,\ldots,q_K\}$. Let $F(p)$ denote the accuracy (or other performance metrics) of the MAS on the task under error rate $p$, then the resilience of the MAS on task $\mathcal{T}$ is defined as:
\begin{equation}
R(\mathcal{G})=\frac{1}{F(0)}\int_{0}^{1}F(p)dp,
\end{equation}
where $F(0)$ represents the original functionality under no perturbation. 
\end{definition}

We visualize this definition in Figure~\ref{subfig:resilience_def}, where resilience is defined as the proportion of area under the curve. 
In this study, we approximate the resilience by testing the accuracy of an MAS under error rates $p=0,0.2,0.4,0.6,0.8,1$ as follows:
\begin{equation}
\label{eqn:resilience}
\begin{aligned}
R(\mathcal{G})=\frac{1}{10F(0)}(&F(0)+2F(0.2)+2F(0.4)\\
&+2F(0.6)+2F(0.8)+F(1)).
\end{aligned}
\end{equation}


Based on the definition above, we further study the factors that affect MAS's resilience. Intuitively, more agents and more communication links between agents will bring more redundancy to the system, resulting in higher error tolerance. 
To verify this, we visualize the relationship between MAS resilience and the number of agents and links in Figure~\ref{fig:motivation}(c)(d), and find that more agents and links do lead to higher resilience, which indicates that we can improve MAS resilience simply by increasing the number of agents and edges. However, since more agents and edges will also bring higher cost, a natural question is whether we can optimize the MAS given the constraints of agent and edge numbers.

\begin{figure}[t]
    \centering
    \includegraphics[width=0.95\linewidth]{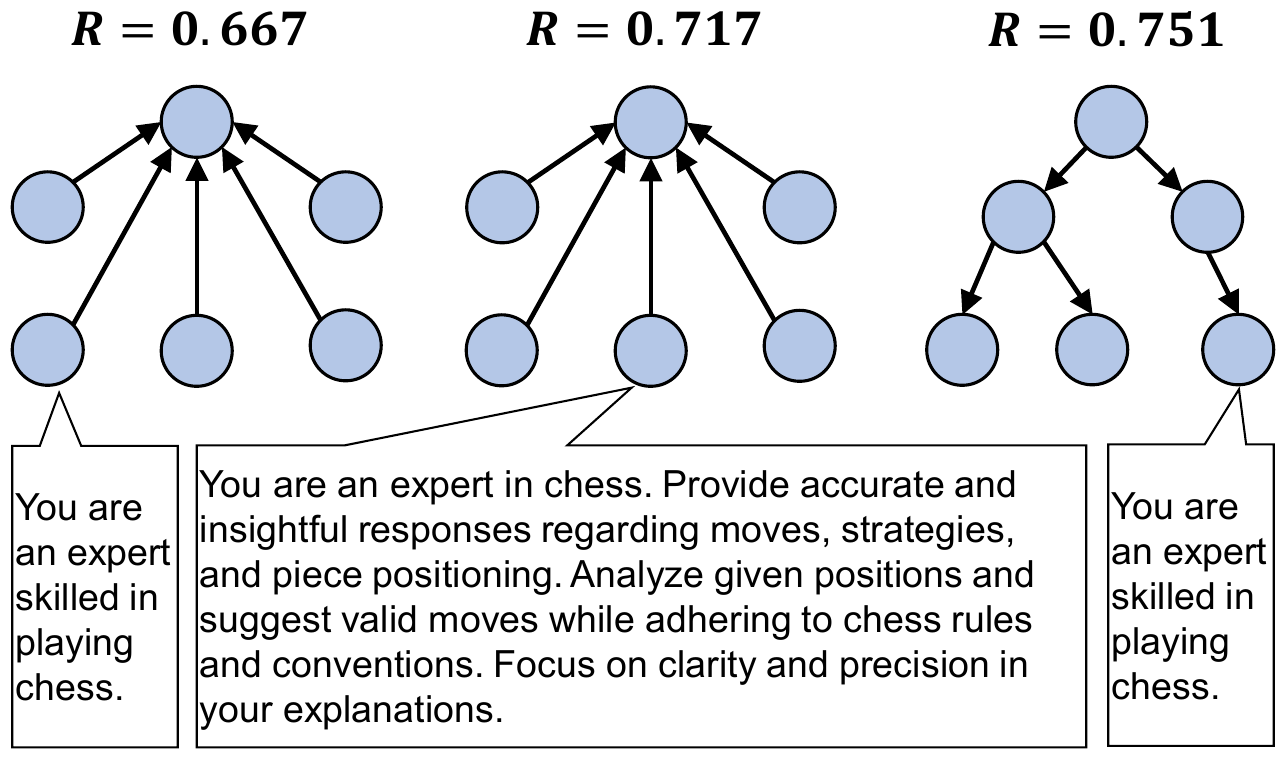}
    \caption{Effect of topology and prompt on MAS resilience.}
    \label{fig:topology_prompt_effect}
\end{figure}

To answer this question, we conduct a case study on MASs with the same number of agents and edges, but with different graph structures and agent prompts. As shown in Figure~\ref{fig:topology_prompt_effect}, we find that when we refine the prompt for each agent (by calling GPT-4o one time), the resilience increases significantly. Moreover, even under the same agent prompts, the hierarchical topology (right) demonstrates higher resilience than the centralized topology (left). These findings suggest both topology and prompt have a high impact on the resilience of MAS.

\subsection{Problem Statement}
Motivated by these, in this work, we aim to design more resilient MAS by optimizing the topology and prompts under the constraints of agent and edge numbers. We formally define the problem as follows:

\begin{definition}[\textbf{MAS Resilience Optimization Problem}]
Given the number of agents $n$ and number of edges $m$, generate an MAS $\mathcal{G}=\{\mathcal{V}, \mathcal{E}, \mathcal{P}\}$ with better resilience on a certain task, i.e.,
\begin{equation}
\begin{aligned}
\arg\max_{\mathcal{E}, \mathcal{P}} \quad & R(\mathcal{G}) \\
\text{s.t.} \quad & |\mathcal{V}|=n, |\mathcal{E}| \leq m.
\end{aligned}
\end{equation}
\end{definition}

\section{Methods}
\label{sec:methods}

\begin{figure*}[h]
    \centering
    \includegraphics[width=0.75\linewidth]{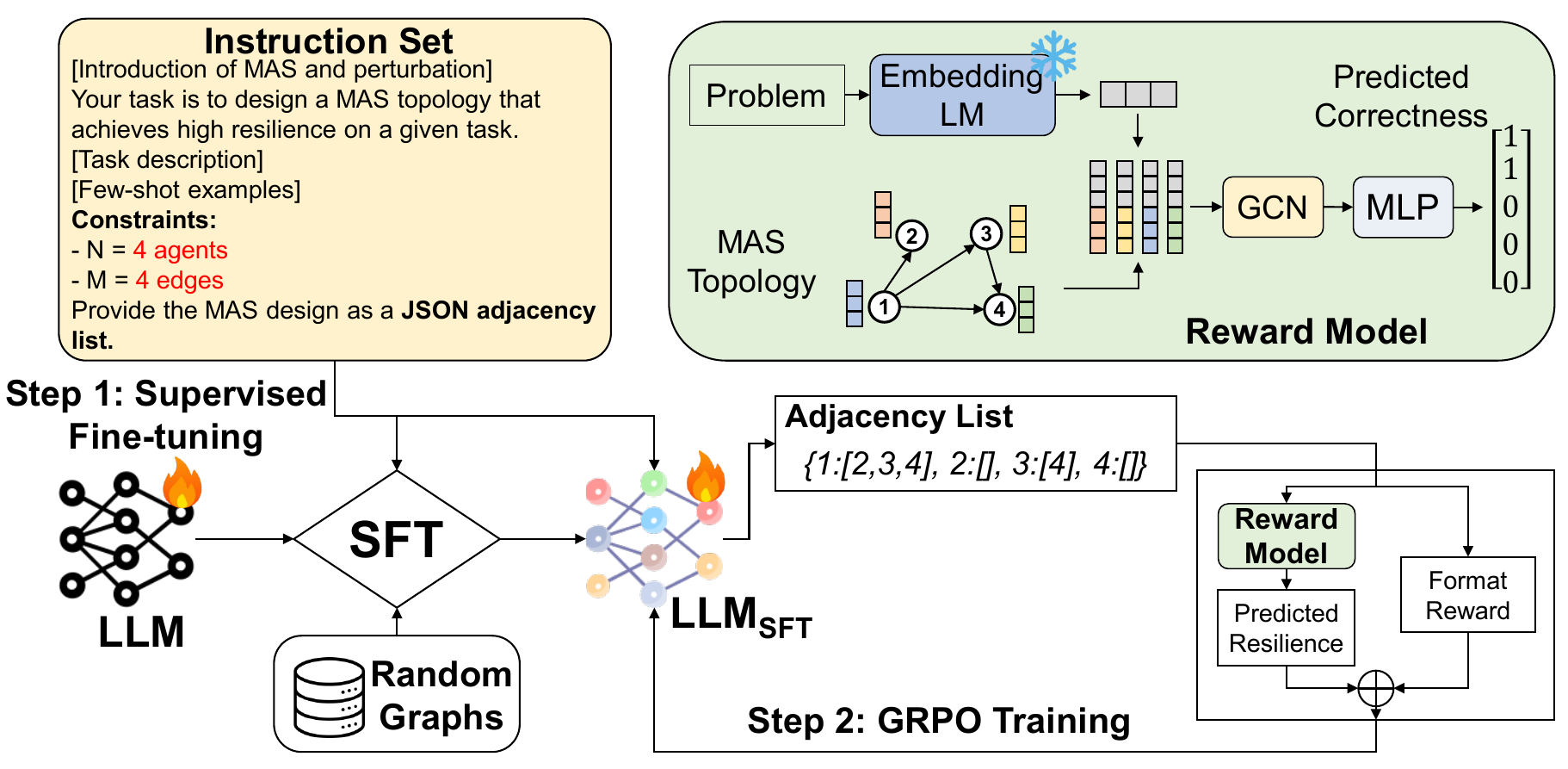}
    \caption{The topology optimization framework in ResMAS.}
    \label{fig:topo_optim}
\end{figure*}

\subsection{Framework Overview}
The optimization targets of the problem include both the topology and agent prompts of MAS, which leads to a vast search space. To reduce the search space, we propose a two-stage framework that first designs the topology of MAS, and then optimizes the prompt with the topology fixed. 

In the first stage, we focus on designing a more resilient MAS topology with identical prompts for all agents. Existing studies have proposed various methods to optimize the topology of MAS to improve its performance~\cite{zhang2024g,zhuge2024gptswarm}. However, these methods are mainly designed for optimizing the accuracy of MAS on certain tasks, and they face two challenges in our resilience optimization problem. First, all existing methods need to evaluate the performance of the proposed MAS on specific tasks, which is highly time-consuming. This issue is further exacerbated in resilience evaluation, as it requires assessing system performance under varying error rates. Second, existing methods typically conduct optimization independently for different tasks, limiting generalization and incurring additional computational cost. To deal with the first challenge, we train a task-aware reward model to predict the resilience of MAS for different tasks. Moreover, inspired by the generalization ability of LLMs, we fine-tune an LLM with reinforcement learning guided by the reward model to serve as the topology generator. The generator can produce resilient topologies for diverse tasks and under various constraints specified via prompts.
In the second stage, we fix the topology from the first stage and optimize the prompts for each agent. Since the prompt of an agent is tightly coupled with its position in the graph and its connections with other agents, we propose a topology-aware prompt optimization method, where each agent's prompt is updated based on the its connections and interactions with neighbors.

\subsection{Topology Optimization}
In the first stage, we aim to optimize the topology of MAS under given node and edge constraints. To make the optimization process generalizable to different tasks and constraints, we train a generator to directly generate the topology instead of searching in the topology space. Inspired by the outstanding instruction following and generalization ability of LLM, we fine-tune a small-sized LLM as the generator (Qwen2.5-7B-Instruct~\cite{qwen2.5}). In this way, a single model can be used for different scenarios by designating the task and constraints in the prompt. The framework of topology optimization is shown in Figure~\ref{fig:topo_optim}.

\subsubsection{Supervised Fine-tuning on Random Graphs}
We first construct the instruction set for training. In each instruction, we provide the introduction of MAS and perturbation, the target task, some few-shot examples, node and edge constraints, and ask the model to output an MAS topology with high resilience in adjacency list format. We vary the agent number from \{10,15,20\}, edge number from 10 to 80, and tasks including MATH, MMLU-Pro, and Chess (details in the dataset section), resulting in 2544 instructions in total. 

In the experiments, we find that it is difficult for LLM to generate valid adjacency lists that adhere to the constraints. Therefore, we first conduct supervised fine-tuning (SFT) to make the model learn the basic required format. Since there is no ground truth for the instructions, we generate random graphs as the labels.  For example, if the constraint in an instruction is “10 agents, 10 edges”, we generate a random graph with 10 nodes and 10 edges as the label for training.

\subsubsection{GRPO Training with Reward Model}
After teaching the model basic format through SFT, we further leverage reinforcement learning to enhance its ability to generate more resilient topologies. However, it is highly time-consuming to evaluate a generated topology since we need to test it under different error rates. For example, calculating the resilience of an MAS with 10 agents on the MATH dataset may take a few hours. To tackle this challenge, we train a reward model to predict the resilience of an MAS on a given task.

Motivated by the outstanding ability of Graph Neural Network to capture characteristics of graphs, we leverage GCN~\cite{kipf2016semi} as the backbone of reward model. Moreover, to enable prediction for different tasks, we use an embedding LM as the task encoder to obtain a low-dimensional embedding for the task, and concatenate it with node features~\cite{zhang2024g}, as shown in Figure~\ref{fig:topo_optim}. Here we initialize the node features with the in-degree and out-degree, and use Sentence-BERT~\cite{reimers2019sentence} as the task encoder.
After information propagation on the graph through two GCN layers, we apply a mean pooling operation to obtain a graph-level embedding, followed by an MLP to generate the final prediction.
Here, we do not directly predict the resilience on the whole dataset, as it would be hard to collect a large amount of training data. Instead, we train the reward model to predict the correctness of each problem in the dataset, i.e., whether the MAS can correctly answer the problem under different error rates. Therefore, the output of the reward model is a 5-dimensional 0-1 vector, representing the correctness under p=0, 0.2, 0.4, 0.6, 0.8. 
For the training data, we use the random graphs from previous analysis (motivation section) to test on three datasets, resulting in 7936 [topology, task, correctness] tuples, which are split into train:test=8:2.  Experiments show that the trained reward model can achieve an accuracy of 0.86 on the test set in predicting the correctness of an MAS on a problem.
In the inference stage, we calculate the predicted resilience $\hat{R}$ on a dataset by aggregating the predicted correctness for all problems in the dataset with Equation~\ref{eqn:resilience}.

After obtaining the reward model that can predict the resilience of any MAS topology on any task, we leverage the GRPO~\cite{shao2024deepseekmath} algorithm to train the LLM to generate more resilient topologies. To ensure the generated topologies satisfy the constraints, we further introduce a format reward that punishes those that are not valid adjacency lists or do not meet the node and edge requirements. The overall reward function is defined as follows:
\begin{equation}
\label{eqn:reward}
    \text{Reward}(\mathcal{G}) = 
    \begin{cases}
    -1, & \text{if wrong format or } |\mathcal{V}| \ne n \\
    -\frac{|\mathcal{E}|-m}{m}, & \text{if } |\mathcal{E}| > m \\
    \hat{R}(\mathcal{G})+|\mathcal{E}|/m,  & \text{if } n = N \text{ and } |\mathcal{E}|\le m
    \end{cases}
\end{equation}
In both the SFT and GRPO training, the model is trained using LoRA~\cite{hu2022lora}.


\subsection{Topology-aware Prompt Optimization}
\begin{figure}[t]
    \centering
    \includegraphics[width=\linewidth]{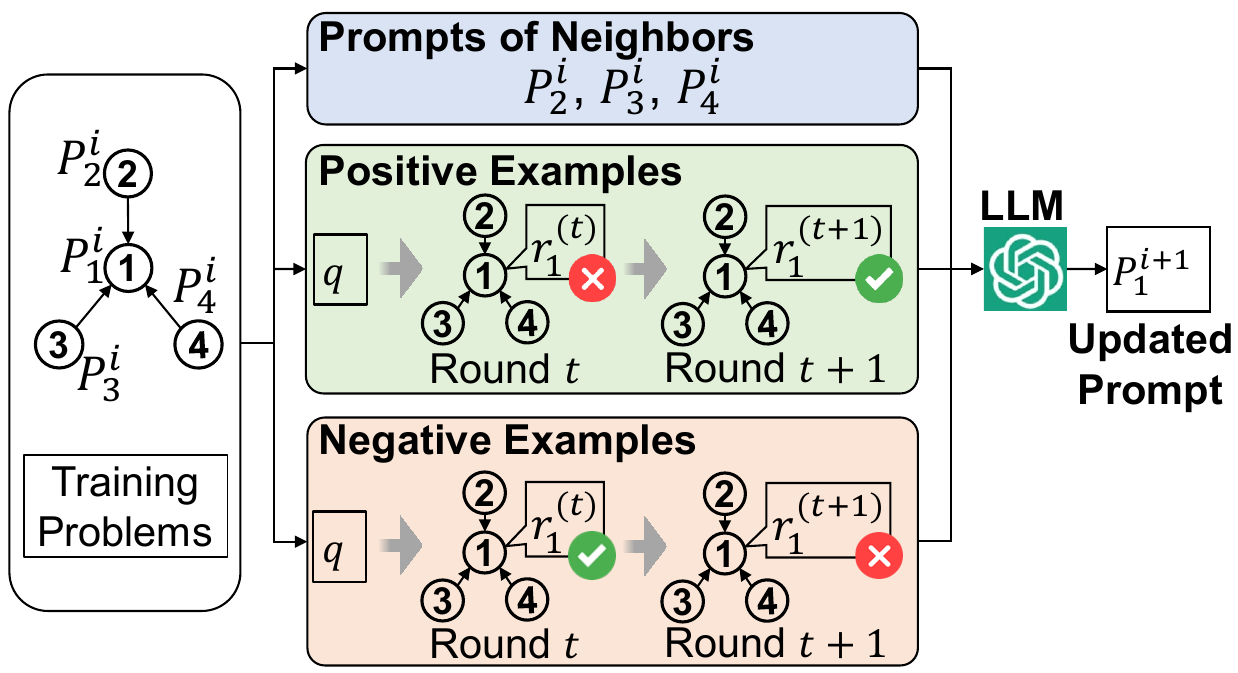}
    \caption{Topology-aware prompt optimization framework in ResMAS.}
    \label{fig:prompt_optim}
\end{figure}

Agents with different positions in the graph and different connections may play different roles in the MAS. Therefore, it is crucial to consider the topology when optimizing prompts. Previous prompt optimization methods for MAS often update the prompt for each agent sequentially~\cite{zhuge2024gptswarm}. Moreover, they are primarily designed for improving the accuracy rather than resilience. In this work, we propose a topology-aware prompt optimization method to address such limitations, as shown in Figure~\ref{fig:prompt_optim}.

The intuition behind our method is that the resilience of MAS is largely affected by the interaction between agents. Simply improving the accuracy of each agent may not bring better resilience, as they may be easily affected by perturbed answers from other agents. Therefore, we hope that the agents can learn when to refer to responses from other agents and when to adhere to their own answers.

Instead of using a simple resilience metric as feedback, we make full use of every problem in the training set. Specifically, we run the MAS on the training set and calculate the correctness of each agent on each round of discussion for each problem. If an agent produces an incorrect response in one round but subsequently corrects its answer in the following round after receiving responses from its neighboring agents, this instance is treated as a positive example. Similarly, we construct negative examples when an agent is misled by responses from neighbors and changes its correct answer to a wrong one. 
Based on the positive examples, negative examples, and the system prompts of neighbors, we leverage a stronger LLM to update the prompt for each agent.
In this way, agents can learn from both the prompts of their neighbors as well as their interaction history with neighbors, enabling topology-aware prompt optimization. 

\section{Experiments}
\label{sec:experiments}

\begin{table*}[h]
\centering
\resizebox{\textwidth}{!}{
\begin{tabular}{c|ccccccccc|ccccccccc}
\toprule
\multirow{3}{*}{\textbf{Method}} & \multicolumn{9}{c|}{\textbf{MATH}}                                                                                                                                                               & \multicolumn{9}{c}{\textbf{MMLU-Pro}}                                                                                                                                                            \\
                                 & \multicolumn{3}{c|}{\textbf{10 Agents}}                               & \multicolumn{3}{c|}{\textbf{15 Agents}}                               & \multicolumn{3}{c|}{\textbf{20 Agents}}          & \multicolumn{3}{c|}{\textbf{10 Agents}}                               & \multicolumn{3}{c|}{\textbf{15 Agents}}                               & \multicolumn{3}{c}{\textbf{20 Agents}}           \\
                                 & \textbf{10E}   & \textbf{20E}   & \multicolumn{1}{c|}{\textbf{30E}}   & \textbf{15E}   & \textbf{30E}   & \multicolumn{1}{c|}{\textbf{45E}}   & \textbf{20E}   & \textbf{40E}   & \textbf{60E}   & \textbf{10E}   & \textbf{20E}   & \multicolumn{1}{c|}{\textbf{30E}}   & \textbf{15E}   & \textbf{30E}   & \multicolumn{1}{c|}{\textbf{45E}}   & \textbf{20E}   & \textbf{40E}   & \textbf{60E}   \\ \hline
G-Designer                       & 0.490          & 0.573          & \multicolumn{1}{c|}{0.531}          & 0.531          & 0.540          & \multicolumn{1}{c|}{0.480}          & 0.546          & 0.572          & 0.558          & 0.456          & 0.498          & \multicolumn{1}{c|}{0.502}          & 0.459          & 0.422          & \multicolumn{1}{c|}{0.454}          & 0.441          & 0.458          & 0.383          \\
OPRO                             & \underline{0.785}    & 0.718          & \multicolumn{1}{c|}{0.752}          & \underline{0.803}    & \underline{0.827}    & \multicolumn{1}{c|}{\underline{0.821}}    & 0.838          & \underline{0.820}    & 0.845          & 0.755          & \underline{0.758}    & \multicolumn{1}{c|}{0.752}          & 0.771          & 0.774          & \multicolumn{1}{c|}{0.812}          & 0.795          & 0.802          & \underline{0.815}    \\
TextGrad                         & 0.756          & 0.746          & \multicolumn{1}{c|}{0.746}          & 0.785          & 0.800          & \multicolumn{1}{c|}{0.820}          & \underline{0.841}    & 0.816          & 0.820          & \underline{0.758}    & 0.731          & \multicolumn{1}{c|}{\underline{0.774}}    & \underline{0.784}    & 0.765          & \multicolumn{1}{c|}{\underline{0.813}}    & 0.782          & 0.806          & 0.810          \\
GPTSwarm                         & 0.752          & \underline{0.760}    & \multicolumn{1}{c|}{\underline{0.786}}    & 0.797          & 0.810          & \multicolumn{1}{c|}{0.786}          & 0.817          & 0.811          & \underline{0.846}    & 0.702          & 0.739          & \multicolumn{1}{c|}{0.702}          & 0.712          & \underline{0.777}    & \multicolumn{1}{c|}{0.789}          & \underline{0.796}    & \underline{0.810}    & 0.808          \\ \hline
ResMAS                             & \textbf{0.813} & \textbf{0.807} & \multicolumn{1}{c|}{\textbf{0.806}} & \textbf{0.847} & \textbf{0.847} & \multicolumn{1}{c|}{\textbf{0.837}} & \textbf{0.849} & \textbf{0.856} & \textbf{0.853} & \textbf{0.766} & \textbf{0.808} & \multicolumn{1}{c|}{\textbf{0.780}} & \textbf{0.808} & \textbf{0.799} & \multicolumn{1}{c|}{\textbf{0.815}} & \textbf{0.817} & \textbf{0.849} & \textbf{0.839} \\ \bottomrule
\end{tabular}

}
\caption{Resilience comparison with baselines on MATH and MMLU-Pro datasets. ``E'' represents the number of edges. Best results are presented in bold, and the second best results are underlined.}
\label{tbl:main_result_math_mmlu}
\end{table*}

\begin{table*}[h]
\centering
\resizebox{0.6\textwidth}{!}{
\begin{tabular}{c|ccccccccc}
\toprule
\multirow{3}{*}{\textbf{Method}} & \multicolumn{9}{c}{\textbf{Chess}}                                                                                                                                                               \\
                                 & \multicolumn{3}{c|}{\textbf{10 Agents}}                               & \multicolumn{3}{c|}{\textbf{15 Agents}}                               & \multicolumn{3}{c}{\textbf{20 Agents}}           \\
                                 & \textbf{10E}   & \textbf{20E}   & \multicolumn{1}{c|}{\textbf{30E}}   & \textbf{15E}   & \textbf{30E}   & \multicolumn{1}{c|}{\textbf{45E}}   & \textbf{20E}   & \textbf{40E}   & \textbf{60E}   \\ \hline
G-Designer                       & 0.707          & \underline{0.731}    & \multicolumn{1}{c|}{\underline{0.767}}    & \underline{0.773}    & 0.717          & \multicolumn{1}{c|}{0.665}          & \underline{0.763}    & \underline{0.776}    & 0.747          \\
OPRO                             & 0.661          & 0.652          & \multicolumn{1}{c|}{0.730}          & 0.764          & 0.753          & \multicolumn{1}{c|}{0.764}          & 0.733          & 0.767          & 0.810          \\
TextGrad                         & 0.762          & 0.700          & \multicolumn{1}{c|}{0.738}          & 0.693          & 0.765          & \multicolumn{1}{c|}{0.750}          & 0.690          & 0.743          & \underline{0.811}    \\
GPTSwarm                         & \underline{0.768}    & 0.712          & \multicolumn{1}{c|}{0.716}          & 0.755          & \underline{0.767}    & \multicolumn{1}{c|}{\underline{0.769}}    & 0.745          & 0.716          & 0.810          \\ \hline
ResMAS                             & \textbf{0.798} & \textbf{0.822} & \multicolumn{1}{c|}{\textbf{0.814}} & \textbf{0.810} & \textbf{0.778} & \multicolumn{1}{c|}{\textbf{0.787}} & \textbf{0.790} & \textbf{0.823} & \textbf{0.823} \\ \bottomrule
\end{tabular}

}
\caption{Resilience comparison with baselines on Chess dataset. ``E'' represents the number of edges. Best results are presented in bold, and the second best results are underlined.}
\label{tbl:main_result_chess}
\end{table*}

\subsection{Experiment Settings}
\subsubsection{Datasets}
\label{sec:datasets}
We evaluate the performance of ResMAS on three datasets including commonsense reasoning, mathematical reasoning, and game, which provides a diverse range of tasks to thoroughly assess our approach.
\begin{itemize}
    \item \textbf{MMLU-Pro}~\cite{wang2024mmlu}: It contains multiple-choice questions from various disciplines with four to ten options, serving as a benchmark to test the commonsense reasoning ability of LLMs. 
    \item \textbf{MATH}~\cite{hendrycks2measuring}: It contains math problems to test the mathematical reasoning ability of LLMs. We choose the hardest level (level-5) in our experiments.
    \item \textbf{Chess}~\cite{srivastava2023beyond}: It is a subset of the BIG-Bench Benchmark containing Chess move validity tasks, where the problems are to provide a valid move of a piece given the chess move history. 
\end{itemize}
During the training of the reward model and topology generator, we only use data from the three datasets above. Moreover, we also evaluate the generalization ability of our method to unseen tasks on HumanEval~\cite{chen2021evaluating} dataset, which measures the code generation ability of LLM.

\subsubsection{Baselines}
We compare our method with three kinds of baselines that optimize topology, prompt, and both of them.

\textbf{Topology optimization baselines.}
\begin{itemize}
    \item \textbf{G-Designer}~\cite{zhang2024g}: It models MAS as a graph, and trains a variational graph auto-encoder to generate the topology for MAS.
\end{itemize}

\textbf{Prompt optimization baselines.}
\begin{itemize}
    \item \textbf{OPRO}~\cite{DBLP:conf/iclr/Yang0LLLZC24}: It uses LLM as optimizer to iteratively generate new prompts based on the performance of existing prompts on the training set.
    \item \textbf{TextGrad}~\cite{yuksekgonul2025optimizing}: It also leverages LLM as optimizer, and optimizes the prompt of the agent by backpropagating feedback, which is similar to training of neural networks. Following the original paper, we use the performance on the training set as the feedback.
\end{itemize}
Since the prompt optimization baselines are designed for a single agent, we use a random graph as the topology of MAS, and use the same optimized prompt for all agents.

\textbf{Topology and prompt optimization baselines.}
\begin{itemize}
    \item \textbf{GPTSwarm}~\cite{zhuge2024gptswarm}: It optimizes the connection between agents through updating the probabilistic distribution of edges based on performance. Moreover, it sequentially optimizes the prompt for each agent in the MAS with LLM as optimizer.
\end{itemize}

We evaluate our method and baselines under different node and edge constraints, including 10-node 10/20/30 edges, 15-node 15/30/45 edges, and 20-node 20/40/60 edges. We use Qwen2.5-32B-Instruct as the backbone model for agents. 

\subsection{Overall Performance}
The overall performance of our method compared with baselines is shown in Table~\ref{tbl:main_result_math_mmlu} and Table~\ref{tbl:main_result_chess}, from which we have the following findings.

First, ResMAS consistently outperforms baselines in terms of resilience on all datasets and across different node and edge constraints, which demonstrates the effectiveness and robustness of our method across various scenarios.

Second, the resilience generally increases with the number of agents and edges, which is consistent with our previous findings (Motivation). However, the performances of different methods vary greatly even under the same constraint, indicating that carefully designing the topology and prompt is essential for a more resilient MAS, and ResMAS manages to generate better MAS across different settings.

Third, we find that G-Designer generally performs the worst among baselines, which indicates that solely optimizing topology is not enough for high resilience. Moreover, GPTSwarm generally performs well because it optimizes both the topology and prompts of MAS. However, it still achieves lower resilience than our methods, which is probably because it fails to consider the topology in the prompt optimization stage, demonstrating the importance of our topology-aware prompt optimization design.

Overall, ResMAS shows robust performance gain across various settings and datasets, showing its effectiveness in jointly optimizing topology and prompts.

\subsection{Optimizing for Accuracy}

\begin{figure}[t]
    \centering
    \includegraphics[width=0.85\linewidth]{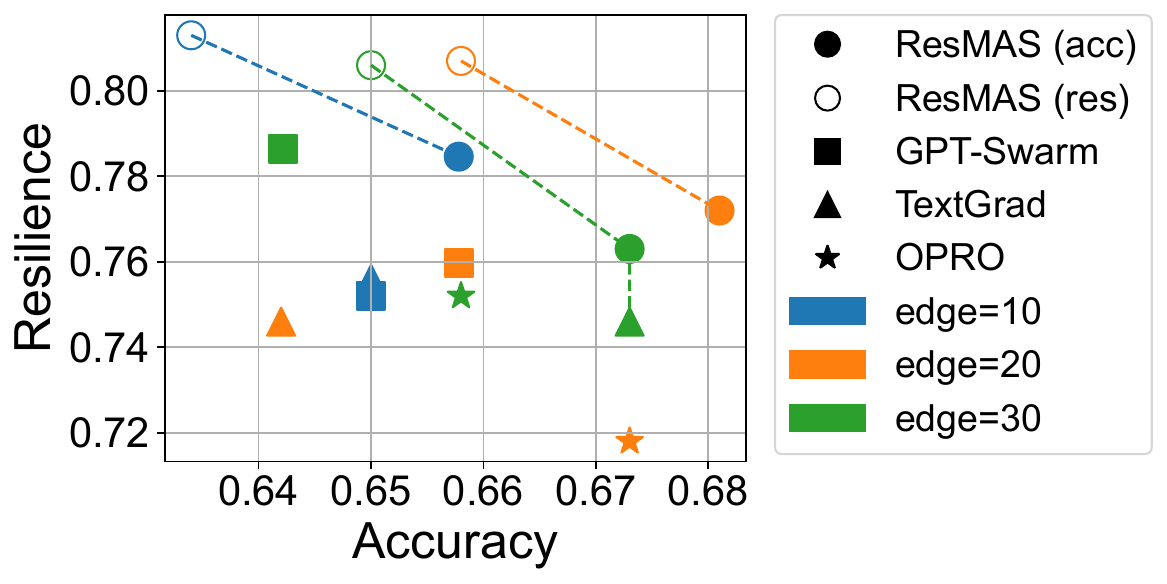}
    \caption{Comparison of accuracy and resilience with baselines. ``ResMAS(acc)'' and ``ResMAS(res)'' represent optimizing accuracy and resilience with our method, respectively. Dashed line represents the Pareto front.}
    \label{fig:acc_res}
\end{figure}
In this section, we show that apart from optimizing resilience, ResMAS can also be used to optimize the performance of MAS. Specifically, since our reward model can predict the correctness of each problem, it can also be directly used for predicting accuracy on the dataset. Therefore, we change the predicted resilience $\hat{R}(\mathcal{G})$ to predicted performance $\hat{P}(\mathcal{G})$ in Equation~\ref{eqn:reward}. Moreover, in the prompt optimization stage, we remove the negative examples since the accuracy is measured under no perturbations. The other procedures are the same as resilience optimization.

We present the results on the MATH dataset under constraints of 10 agents and 10/20/30 edges in Figure~\ref{fig:acc_res}. The edge constraints are denoted by colors, and different methods are denoted by shapes. For comparison, we also present the original resilience optimization results of our method. It can be observed that when changing the optimization goal to accuracy, the accuracy of generated MAS increases with a slight drop in resilience. Moreover, our method consistently achieves the Pareto front (dashed line) under different constraints. Such results demonstrate that ResMAS can optimize both the resilience and accuracy of MAS effectively.

\subsection{Generalization Ability}
To evaluate the generalization ability of ResMAS, we apply our method to new tasks or new models without re-training the topology generator, and compare with topology optimization baselines.

\begin{figure}[t]
\centering
\includegraphics[width=\linewidth]{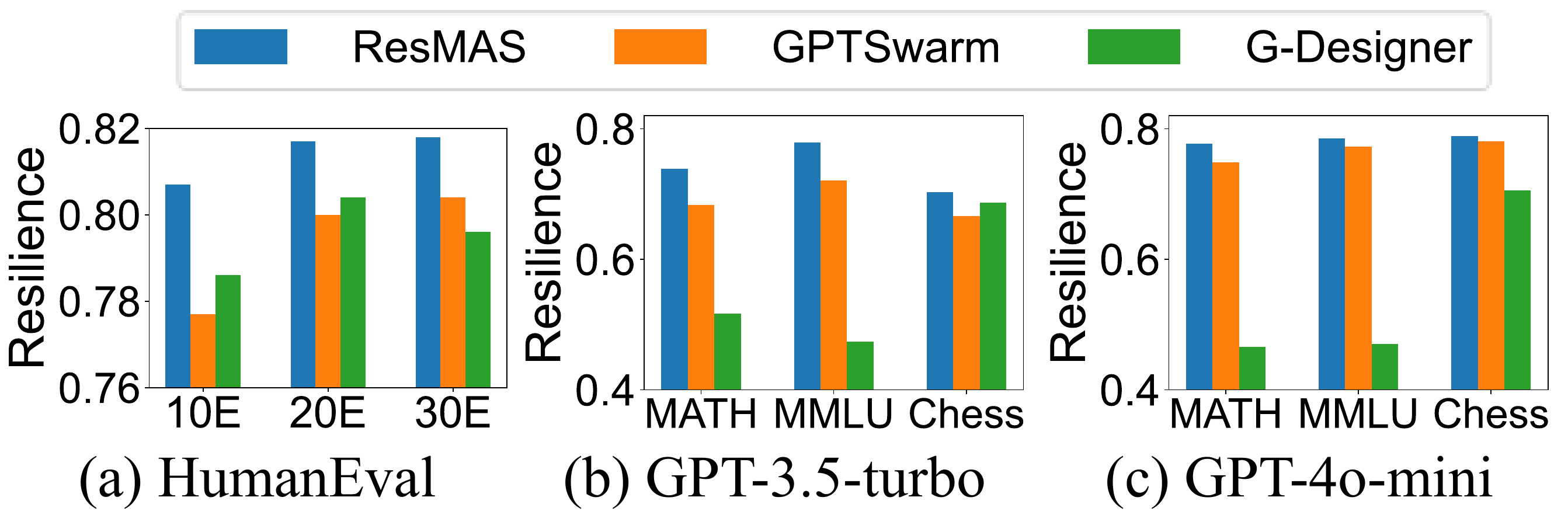}
\caption{Generalization ability of ResMAS on new task (a) and new models (b) (c). ``E'' means the number of edges.}
\label{fig:generalization}
\end{figure}

For cross-task generalization, we use our topology generator to directly generate MAS topology for the HumanEval dataset, which is unseen in the training data. Then we use the training set of HumanEval to optimize the prompts. As shown in Figure~\ref{fig:generalization}(a), our method achieves better performance than the baselines, indicating that our method can generalize well to new tasks.
Furthermore, we transfer our method to agents with different backbone LLMs, i.e., GPT-3.5-turbo and GPT-4o-mini. Here we use the same topology but re-optimize the prompts for each model. As shown in Figure~\ref{fig:generalization}(b) and (c), our method consistently performs better than baselines on different backbone models and different tasks, indicating that our generated MAS topology is generalizable to agents with different LLM backbones.


\begin{figure}[t]
\centering
\includegraphics[width=\linewidth]{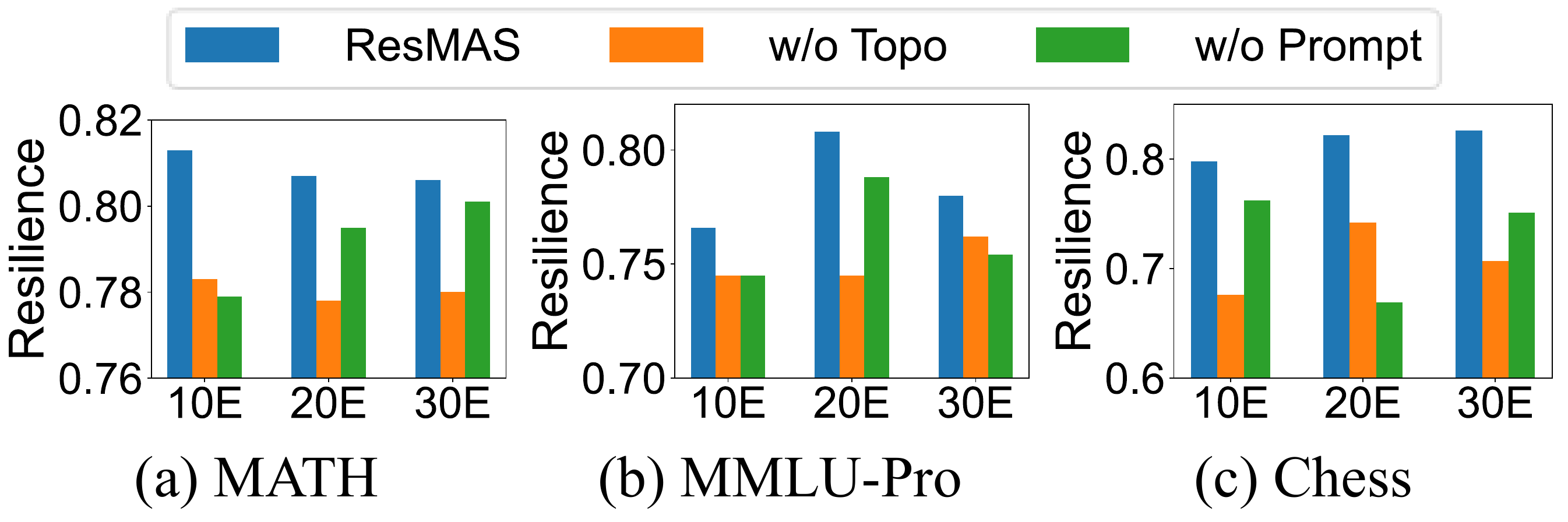}
\caption{Resilience comparison of ResMAS without topology or prompt optimization. ``E'' means the number of edges.}
\label{fig:ablation}
\end{figure}

\subsection{Ablation Study}
To validate the effectiveness of optimizing topology and prompt, we conduct an ablation study on the two stages. For the ablation of topology optimization, we use a random graph as the MAS topology and then optimize the prompt. As for prompt optimization, we use the identical initial prompt for all agents. The results on three datasets are shown in Figure~\ref{fig:ablation}, from which we find that the resilience drops in all cases when removing either stage, demonstrating the effectiveness of them.


\subsection{Case Study}
\begin{figure}[t]
    \centering
    \includegraphics[width=0.98\linewidth]{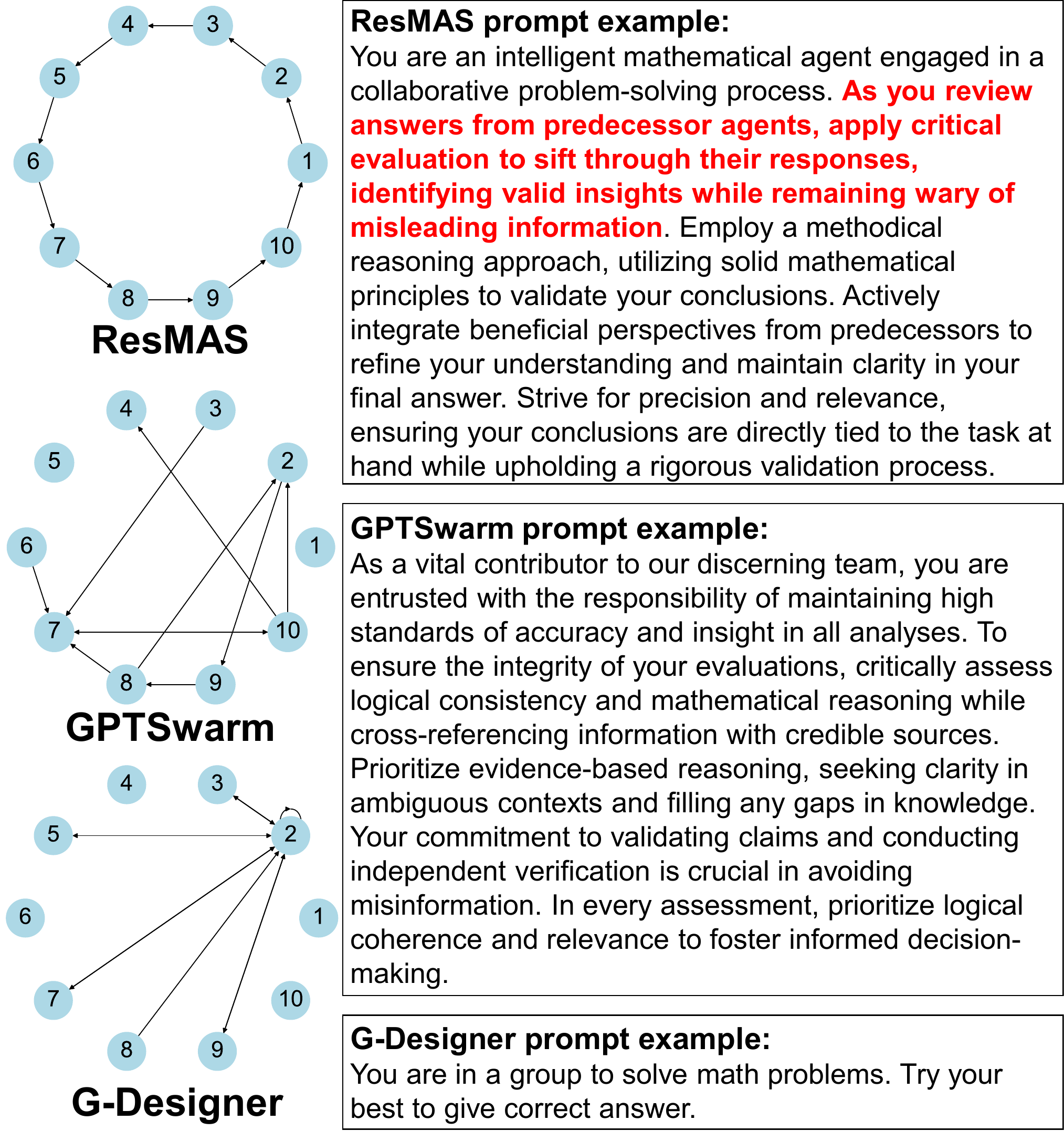}
    \caption{Topologies and prompts of MAS generated by different methods.}
    \label{fig:case}
\end{figure}

To intuitively understand how ResMAS achieves better resilience, we compare our topology and prompts with baselines (10 agents, 10 edges, MATH dataset) in Figure~\ref{fig:case}.
We find that our method generates a decentralized graph where each node has the same degree. In comparison, G-Designer tends to generate a centralized graph, where multiple nodes are connected to node 2. Moreover, both GPTSwarm and G-Designer fail to generate a connected graph, where the isolated agents may compromise the resilience of MAS.
As for prompts, since G-Designer does not optimize the prompts, all agents share the same initial prompt. GPTSwarm and our method both conduct prompt optimization, while only our method considers the topology and interaction between agents. As a result, our prompt explicitly instructs the agent to pay attention to potential perturbations from neighbors, which further enhances the resilience of the system.

\section{Conclusion}
\label{sec:conclusion}

In this paper, we conduct a comprehensive study on the resilience of MAS under random agent failures, and propose a two-stage framework ResMAS to optimize the topology and prompts of MAS to improve resilience. 
Experiments on diverse benchmarks and settings demonstrate the effectiveness, versatility, and generalization ability of our approach.
In the future, one potential direction is to jointly optimize the topology and prompts rather than treating them in separate stages. Another direction is to extend our framework to heterogeneous MAS composed of agents with different backbone models or capabilities, which more closely mirrors real-world deployment scenarios.

\section{Acknowledgments}
This work was supported in part by the National Natural Science Foundation of China under Grant 62472241 and Grant 23IAA02114; in part by Tsinghua-Toyota Joint Research Institute Inter-disciplinary Program.

\bibliography{aaai2026}

@inproceedings{zhang2024psysafe,
  title={PsySafe: A Comprehensive Framework for Psychological-based Attack, Defense, and Evaluation of Multi-agent System Safety},
  author={Zhang, Zaibin and Zhang, Yongting and Li, Lijun and Shao, Jing and Gao, Hongzhi and Qiao, Yu and Wang, Lijun and Lu, Huchuan and Zhao, Feng},
  booktitle={Proceedings of the 62nd Annual Meeting of the Association for Computational Linguistics (Volume 1: Long Papers)},
  pages={15202--15231},
  year={2024}
}

@article{wang2025g,
  title={G-safeguard: A topology-guided security lens and treatment on llm-based multi-agent systems},
  author={Wang, Shilong and Zhang, Guibin and Yu, Miao and Wan, Guancheng and Meng, Fanci and Guo, Chongye and Wang, Kun and Wang, Yang},
  journal={arXiv preprint arXiv:2502.11127},
  year={2025}
}

@article{ju2024flooding,
  title={Flooding spread of manipulated knowledge in llm-based multi-agent communities},
  author={Ju, Tianjie and Wang, Yiting and Ma, Xinbei and Cheng, Pengzhou and Zhao, Haodong and Wang, Yulong and Liu, Lifeng and Xie, Jian and Zhang, Zhuosheng and Liu, Gongshen},
  journal={arXiv preprint arXiv:2407.07791},
  year={2024}
}

@inproceedings{gu2024agent,
  title={Agent Smith: a single image can jailbreak one million multimodal LLM agents exponentially fast},
  author={Gu, Xiangming and Zheng, Xiaosen and Pang, Tianyu and Du, Chao and Liu, Qian and Wang, Ye and Jiang, Jing and Lin, Min},
  booktitle={Proceedings of the 41st International Conference on Machine Learning},
  pages={16647--16672},
  year={2024}
}

@article{tian2023evil,
  title={Evil geniuses: Delving into the safety of llm-based agents},
  author={Tian, Yu and Yang, Xiao and Zhang, Jingyuan and Dong, Yinpeng and Su, Hang},
  journal={arXiv preprint arXiv:2311.11855},
  year={2023}
}

@inproceedings{huangresilience,
  title={On the Resilience of LLM-Based Multi-Agent Collaboration with Faulty Agents},
  author={Huang, Jen-tse and Zhou, Jiaxu and Jin, Tailin and Zhou, Xuhui and Chen, Zixi and Wang, Wenxuan and Yuan, Youliang and Lyu, Michael and Sap, Maarten},
  booktitle={Forty-second International Conference on Machine Learning},
  year={2025}
}

@article{yu2024netsafe,
  title={Netsafe: Exploring the topological safety of multi-agent networks},
  author={Yu, Miao and Wang, Shilong and Zhang, Guibin and Mao, Junyuan and Yin, Chenlong and Liu, Qijiong and Wen, Qingsong and Wang, Kun and Wang, Yang},
  journal={arXiv preprint arXiv:2410.15686},
  year={2024}
}

@article{lee2024prompt,
  title={Prompt infection: Llm-to-llm prompt injection within multi-agent systems},
  author={Lee, Donghyun and Tiwari, Mo},
  journal={arXiv preprint arXiv:2410.07283},
  year={2024}
}

@inproceedings{qian2024chatdev,
  title={ChatDev: Communicative Agents for Software Development},
  author={Qian, Chen and Liu, Wei and Liu, Hongzhang and Chen, Nuo and Dang, Yufan and Li, Jiahao and Yang, Cheng and Chen, Weize and Su, Yusheng and Cong, Xin and others},
  booktitle={Proceedings of the 62nd Annual Meeting of the Association for Computational Linguistics (Volume 1: Long Papers)},
  pages={15174--15186},
  year={2024}
}

@inproceedings{hong2023metagpt,
  title={MetaGPT: Meta programming for a multi-agent collaborative framework},
  author={Hong, Sirui and Zhuge, Mingchen and Chen, Jonathan and Zheng, Xiawu and Cheng, Yuheng and Wang, Jinlin and Zhang, Ceyao and Wang, Zili and Yau, Steven Ka Shing and Lin, Zijuan and others},
  booktitle={The Twelfth International Conference on Learning Representations},
  year={2023}
}

@inproceedings{zhuge2024gptswarm,
  title={Gptswarm: Language agents as optimizable graphs},
  author={Zhuge, Mingchen and Wang, Wenyi and Kirsch, Louis and Faccio, Francesco and Khizbullin, Dmitrii and Schmidhuber, J{\"u}rgen},
  booktitle={Forty-first International Conference on Machine Learning},
  year={2024}
}

@article{zhang2024g,
  title={G-designer: Architecting multi-agent communication topologies via graph neural networks},
  author={Zhang, Guibin and Yue, Yanwei and Sun, Xiangguo and Wan, Guancheng and Yu, Miao and Fang, Junfeng and Wang, Kun and Chen, Tianlong and Cheng, Dawei},
  journal={arXiv preprint arXiv:2410.11782},
  year={2024}
}

@article{yuksekgonul2025optimizing,
  title={Optimizing generative AI by backpropagating language model feedback},
  author={Yuksekgonul, Mert and Bianchi, Federico and Boen, Joseph and Liu, Sheng and Lu, Pan and Huang, Zhi and Guestrin, Carlos and Zou, James},
  journal={Nature},
  volume={639},
  number={8055},
  pages={609--616},
  year={2025},
  publisher={Nature Publishing Group}
}

@article{wang2024mmlu,
  title={Mmlu-pro: A more robust and challenging multi-task language understanding benchmark},
  author={Wang, Yubo and Ma, Xueguang and Zhang, Ge and Ni, Yuansheng and Chandra, Abhranil and Guo, Shiguang and Ren, Weiming and Arulraj, Aaran and He, Xuan and Jiang, Ziyan and others},
  journal={Advances in Neural Information Processing Systems},
  volume={37},
  pages={95266--95290},
  year={2024}
}

@inproceedings{hendrycks2measuring,
  author       = {Dan Hendrycks and
                  Collin Burns and
                  Saurav Kadavath and
                  Akul Arora and
                  Steven Basart and
                  Eric Tang and
                  Dawn Song and
                  Jacob Steinhardt},
  editor       = {Joaquin Vanschoren and
                  Sai{-}Kit Yeung},
  title        = {Measuring Mathematical Problem Solving With the {MATH} Dataset},
  booktitle    = {Proceedings of the Neural Information Processing Systems Track on
                  Datasets and Benchmarks 1, NeurIPS Datasets and Benchmarks 2021, December
                  2021, virtual},
  year         = {2021},
  url          = {https://datasets-benchmarks-proceedings.neurips.cc/paper/2021/hash/be83ab3ecd0db773eb2dc1b0a17836a1-Abstract-round2.html},
  bibsource    = {dblp computer science bibliography, https://dblp.org}
}

@article{srivastava2023beyond,
  title={Beyond the Imitation Game: Quantifying and extrapolating the capabilities of language models},
  author={Srivastava, Aarohi and Rastogi, Abhinav and Rao, Abhishek and Shoeb, Abu Awal Md and Abid, Abubakar and Fisch, Adam and Brown, Adam R and Santoro, Adam and Gupta, Aditya and Garriga-Alonso, Adri{\`a} and others},
  journal={Transactions on Machine Learning Research},
  year={2023}
}

@article{chen2021evaluating,
  title={Evaluating large language models trained on code},
  author={Chen, Mark and Tworek, Jerry and Jun, Heewoo and Yuan, Qiming and Pinto, Henrique Ponde De Oliveira and Kaplan, Jared and Edwards, Harri and Burda, Yuri and Joseph, Nicholas and Brockman, Greg and others},
  journal={arXiv preprint arXiv:2107.03374},
  year={2021}
}

@inproceedings{wang2025mixtureofagents,
title={Mixture-of-Agents Enhances Large Language Model Capabilities},
author={Junlin Wang and Jue WANG and Ben Athiwaratkun and Ce Zhang and James Zou},
booktitle={The Thirteenth International Conference on Learning Representations},
year={2025},
url={https://openreview.net/forum?id=h0ZfDIrj7T}
}

@inproceedings{du2023improving,
  title={Improving factuality and reasoning in language models through multiagent debate},
  author={Du, Yilun and Li, Shuang and Torralba, Antonio and Tenenbaum, Joshua B and Mordatch, Igor},
  booktitle={Forty-first International Conference on Machine Learning},
  year={2023}
}

@inproceedings{li2023camel,
    title={{CAMEL}: Communicative Agents for ''Mind'' Exploration of Large Language Model Society},
    author={Guohao Li and Hasan Abed Al Kader Hammoud and Hani Itani and Dmitrii Khizbullin and Bernard Ghanem},
    booktitle={Thirty-seventh Conference on Neural Information Processing Systems},
    year={2023},
    url={https://openreview.net/forum?id=3IyL2XWDkG}
}

@inproceedings{xiao2023chain,
  title={Chain-of-Experts: When LLMs Meet Complex Operations Research Problems},
  author={Xiao, Ziyang and Zhang, Dongxiang and Wu, Yangjun and Xu, Lilin and Wang, Yuan Jessica and Han, Xiongwei and Fu, Xiaojin and Zhong, Tao and Zeng, Jia and Song, Mingli and others},
  booktitle={The Twelfth International Conference on Learning Representations},
  year={2023}
}

@article{cohen2000resilience,
  title={Resilience of the internet to random breakdowns},
  author={Cohen, Reuven and Erez, Keren and Ben-Avraham, Daniel and Havlin, Shlomo},
  journal={Physical review letters},
  volume={85},
  number={21},
  pages={4626},
  year={2000},
  publisher={APS}
}

@article{gao2016universal,
  title={Universal resilience patterns in complex networks},
  author={Gao, Jianxi and Barzel, Baruch and Barab{\'a}si, Albert-L{\'a}szl{\'o}},
  journal={Nature},
  volume={530},
  number={7590},
  pages={307--312},
  year={2016},
  publisher={Nature Publishing Group UK London}
}

@article{ju2025investigating,
  title={Investigating the Adaptive Robustness with Knowledge Conflicts in LLM-based Multi-Agent Systems},
  author={Ju, Tianjie and Wang, Bowen and Fei, Hao and Lee, Mong-Li and Hsu, Wynne and Li, Yun and Wang, Qianren and Cheng, Pengzhou and Wu, Zongru and Zhang, Zhuosheng and others},
  journal={arXiv preprint arXiv:2502.15153},
  year={2025}
}

@article{shao2024deepseekmath,
  title={Deepseekmath: Pushing the limits of mathematical reasoning in open language models},
  author={Shao, Zhihong and Wang, Peiyi and Zhu, Qihao and Xu, Runxin and Song, Junxiao and Bi, Xiao and Zhang, Haowei and Zhang, Mingchuan and Li, YK and Wu, Yang and others},
  journal={arXiv preprint arXiv:2402.03300},
  year={2024}
}

@misc{qwen2.5,
    title = {Qwen2.5: A Party of Foundation Models},
    url = {https://qwenlm.github.io/blog/qwen2.5/},
    author = {Qwen Team},
    month = {September},
    year = {2024}
}

@article{kipf2016semi,
  title={Semi-Supervised Classification with Graph Convolutional Networks},
  author={Kipf, TN},
  journal={arXiv preprint arXiv:1609.02907},
  year={2016}
}

@inproceedings{reimers2019sentence,
  title={Sentence-BERT: Sentence Embeddings using Siamese BERT-Networks},
  author={Reimers, Nils and Gurevych, Iryna},
  booktitle={Proceedings of the 2019 Conference on Empirical Methods in Natural Language Processing and the 9th International Joint Conference on Natural Language Processing (EMNLP-IJCNLP)},
  pages={3982--3992},
  year={2019}
}

@inproceedings{hu2022lora,
  author       = {Edward J. Hu and
                  Yelong Shen and
                  Phillip Wallis and
                  Zeyuan Allen{-}Zhu and
                  Yuanzhi Li and
                  Shean Wang and
                  Lu Wang and
                  Weizhu Chen},
  title        = {LoRA: Low-Rank Adaptation of Large Language Models},
  booktitle    = {The Tenth International Conference on Learning Representations, {ICLR}
                  2022, Virtual Event, April 25-29, 2022},
  publisher    = {OpenReview.net},
  year         = {2022},
  url          = {https://openreview.net/forum?id=nZeVKeeFYf9},
  bibsource    = {dblp computer science bibliography, https://dblp.org}
}

@inproceedings{DBLP:conf/icml/Du00TM24,
  author       = {Yilun Du and
                  Shuang Li and
                  Antonio Torralba and
                  Joshua B. Tenenbaum and
                  Igor Mordatch},
  title        = {Improving Factuality and Reasoning in Language Models through Multiagent
                  Debate},
  booktitle    = {Forty-first International Conference on Machine Learning, {ICML} 2024,
                  Vienna, Austria, July 21-27, 2024},
  publisher    = {OpenReview.net},
  year         = {2024},
  url          = {https://openreview.net/forum?id=zj7YuTE4t8},
  bibsource    = {dblp computer science bibliography, https://dblp.org}
}

@inproceedings{chen2023agentverse,
  title={Agentverse: Facilitating multi-agent collaboration and exploring emergent behaviors},
  author={Chen, Weize and Su, Yusheng and Zuo, Jingwei and Yang, Cheng and Yuan, Chenfei and Chan, Chi-Min and Yu, Heyang and Lu, Yaxi and Hung, Yi-Hsin and Qian, Chen and others},
  booktitle={The Twelfth International Conference on Learning Representations},
  year={2023}
}

@inproceedings{DBLP:conf/iclr/Yang0LLLZC24,
  author       = {Chengrun Yang and
                  Xuezhi Wang and
                  Yifeng Lu and
                  Hanxiao Liu and
                  Quoc V. Le and
                  Denny Zhou and
                  Xinyun Chen},
  title        = {Large Language Models as Optimizers},
  booktitle    = {The Twelfth International Conference on Learning Representations,
                  {ICLR} 2024, Vienna, Austria, May 7-11, 2024},
  publisher    = {OpenReview.net},
  year         = {2024},
  url          = {https://openreview.net/forum?id=Bb4VGOWELI},
  bibsource    = {dblp computer science bibliography, https://dblp.org}
}

@article{ding2024comprehensive,
  title={A Comprehensive Survey on Artificial Intelligence for Complex Network: Potential, Methodology and Application},
  author={Ding, Jingtao and Liu, Chang and Zheng, Yu and Zhang, Yunke and Yu, Zihan and Li, Ruikun and Chen, Hongyi and Piao, Jinghua and Wang, Huandong and Liu, Jiazhen and others},
  journal={arXiv preprint arXiv:2402.16887},
  year={2024}
}

\clearpage

\section{Appendix}

\subsection{Implementation Details}
\subsubsection{Details of Our Method}
For the reward model, we train the model for 200 epochs with a learning rate of 0.001. We adopt an early stopping mechanism and choose the epoch with the highest accuracy on the training set.
For GRPO training, we set the learning rate to 1e-4 and train for 5 epochs.
For prompt optimization, we choose 32 questions from each dataset as the training set, and set the batch size to 4, i.e., we test the MAS on 4 questions and use the results to update the prompts once. We use gpt-4o-mini as the prompt optimizer for our method and all baselines.
Our code is open-source\footnote{\url{https://github.com/tsinghua-fib-lab/ResMAS}} for reproducibility.

\subsubsection{Details of Baselines}
For topology optimization baselines, including G-Designer and GPTSwarm, the models output a probabilistic adjacency matrix. We apply appropriate thresholding to ensure that the generated graphs satisfy the edge constraints. For example, when the number of edges is set to 10, we retain the 10 edges with the highest probabilities in the matrix.

\subsubsection{Computational Cost}
Experiments are conducted on NVIDIA A100 80G GPUs. The computational cost of our method is as follows.
\begin{itemize}
    \item \textbf{Reward model}: 1 GPU, 2 min.
    \item \textbf{SFT}: 1 GPU, 25 min.
    \item \textbf{GRPO}: 8 GPUs, 12 h.
    \item \textbf{Prompt optimization}: The maximum computational cost for each dataset (MAS with 20 agents and 60 edges) is as follows: MATH requires 22 minutes and 4.1M tokens (\$0.9); MMLU-Pro requires 12 minutes and 2.8M tokens (\$0.6); Chess requires 4 minutes and 0.7M tokens (\$0.1).
\end{itemize}
Note that the reward model, SFT, and GRPO are trained only once. The prompt optimization is performed separately for each dataset and for each configuration of node and edge constraints.

\subsection{Prompts}
\begin{figure*}[h]
    \centering
    \includegraphics[width=.9\linewidth]{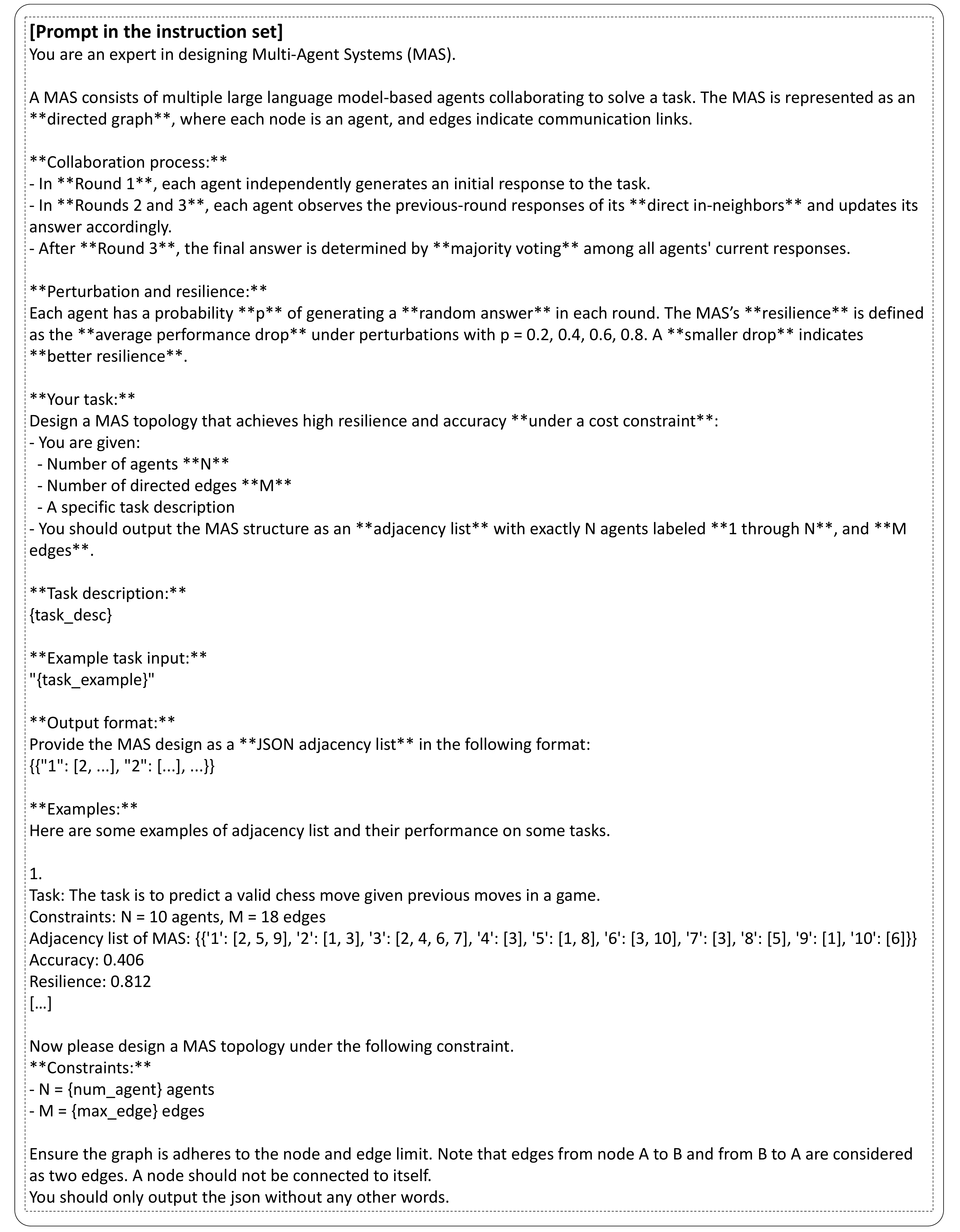}
    \caption{Prompt in the instruction set.}
\label{fig:prompt_instruction}
\end{figure*}

\begin{figure*}[h]
    \centering
    \includegraphics[width=.9\linewidth]{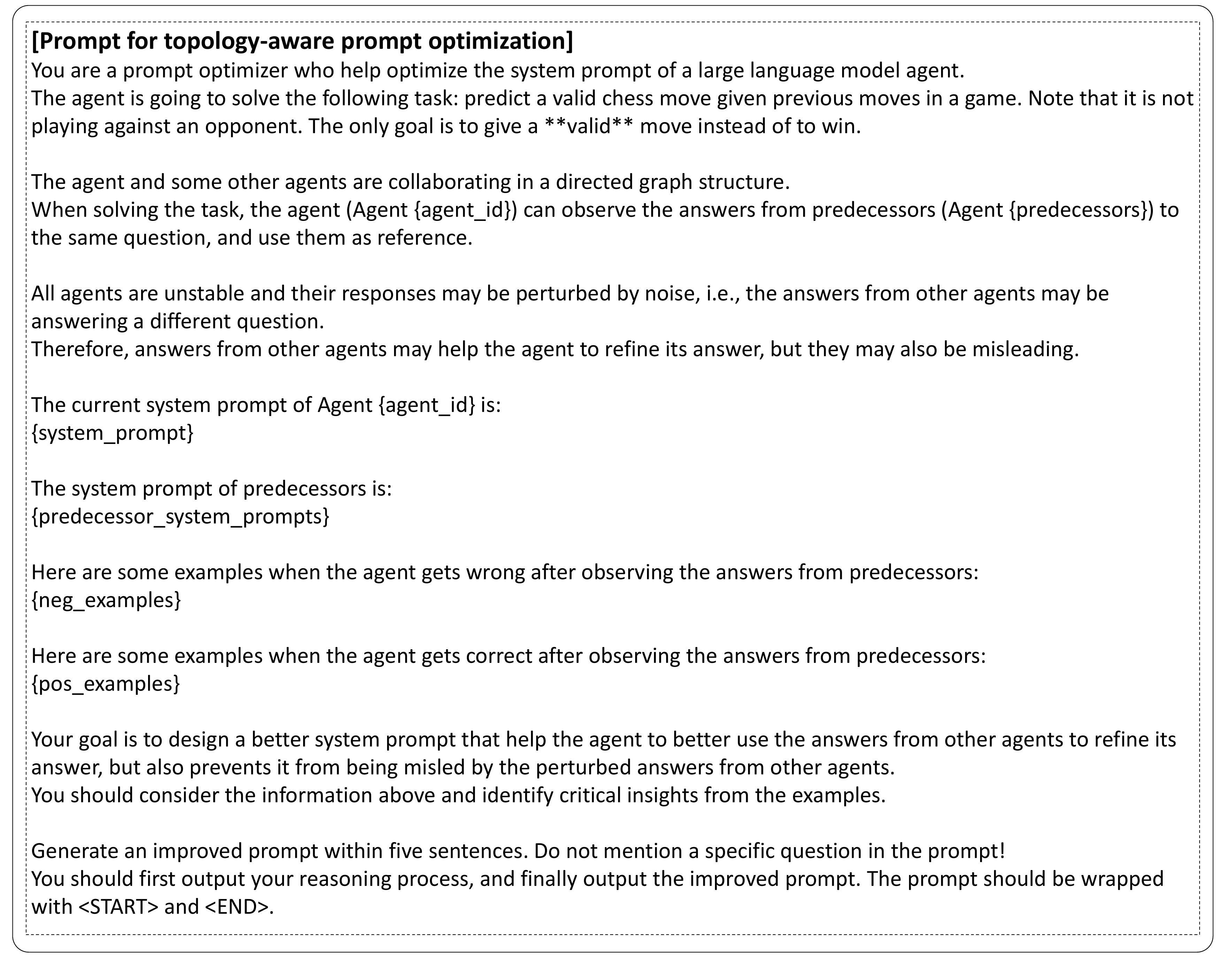}
    \caption{Prompt for topology-aware prompt optimization.}
\label{fig:prompt_optimization_prompt}
\end{figure*}

The prompt in the instruction set for SFT and GRPO training is shown in Figure~\ref{fig:prompt_instruction}.
The prompt for topology-aware prompt optimization is shown in Figure~\ref{fig:prompt_optimization_prompt}

\end{document}